\documentclass[conference]{IEEEtran}
\usepackage{cite}
\usepackage{amsmath,amssymb,amsfonts}
\usepackage{graphicx}
\usepackage{textcomp}
\usepackage{xcolor}
\usepackage{hyperref}
\usepackage{booktabs}
\def\BibTeX{{\rm B\kern-.05em{\sc i\kern-.025em b}\kern-.08em
    T\kern-.1667em\lower.7ex\hbox{E}\kern-.125emX}}

\begin{document}

\title{Beyond AHI: An Interpretable Causal-Discovery--Guided Framework for Sleep Recovery in Connected Health}
\author{
Saba A. Farahani$^*$\quad
Elahe Khatibi\quad
Manoj Vishwanath\quad
Amir M. Rahmani\quad
Hung Cao\\[4pt]
\small University of California, Irvine, CA, USA\\
\small \texttt{\{fazizaba, ekhatibi, manojv, a.rahmani, hungcao\}@uci.edu}
}
\maketitle
\begin{abstract}
Objective sleep assessment relies on polysomnography (PSG), 
yet clinical impact is often better reflected in 
patient-reported outcomes (PROs) such as sleepiness and 
fatigue. Existing summary indices, including the 
Apnea--Hypopnea Index (AHI), provide limited insight into 
the multidomain physiology underlying functional recovery. 
We propose an interpretable, causal-discovery--guided 
framework for deriving a hierarchical Sleep Recovery Score 
(SRS) from multimodal PSG. Using two large population cohorts (MESA: $n=1{,}540$; MrOS: $n=825$), we apply directed acyclic graph 
(DAG) learning to identify candidate physiological drivers 
spanning respiratory burden, hypoxic burden, sleep 
fragmentation, sleep architecture, and autonomic regulation. 
Although derived from clinical PSG, these domains map 
naturally to sensing streams increasingly available in 
connected health technologies, including wearable ECG, 
oximetry, and sleep-stage estimation devices. To preserve 
mechanistic plausibility, we introduce a two-stage screening 
process that combines physiology-based constraints with 
constrained LLM-assisted auditing to identify and remove 
structural confounders and construct-overlapping variables. Across cohorts, these five domains emerge as recurrent physiological domains associated with recovery, and the resulting SRS shows up to 3.4$\times$ stronger alignment with perceived recovery than AHI. By linking multimodal sleep physiology to patient-centered outcomes through an interpretable, bias-aware, and domain structured framework, this work provides a practical foundation for recovery modeling across both clinical sleep studies and emerging smart and connected health settings.
\end{abstract}

\noindent\textbf{Code Availability.}
Code, prompt templates, and classification criteria are available at
\href{https://github.com/elakhatibi/SRS-causal-discovery}
{GitHub}.

\begin{figure*}[t]
  \centering
  \includegraphics[width=0.98\textwidth]{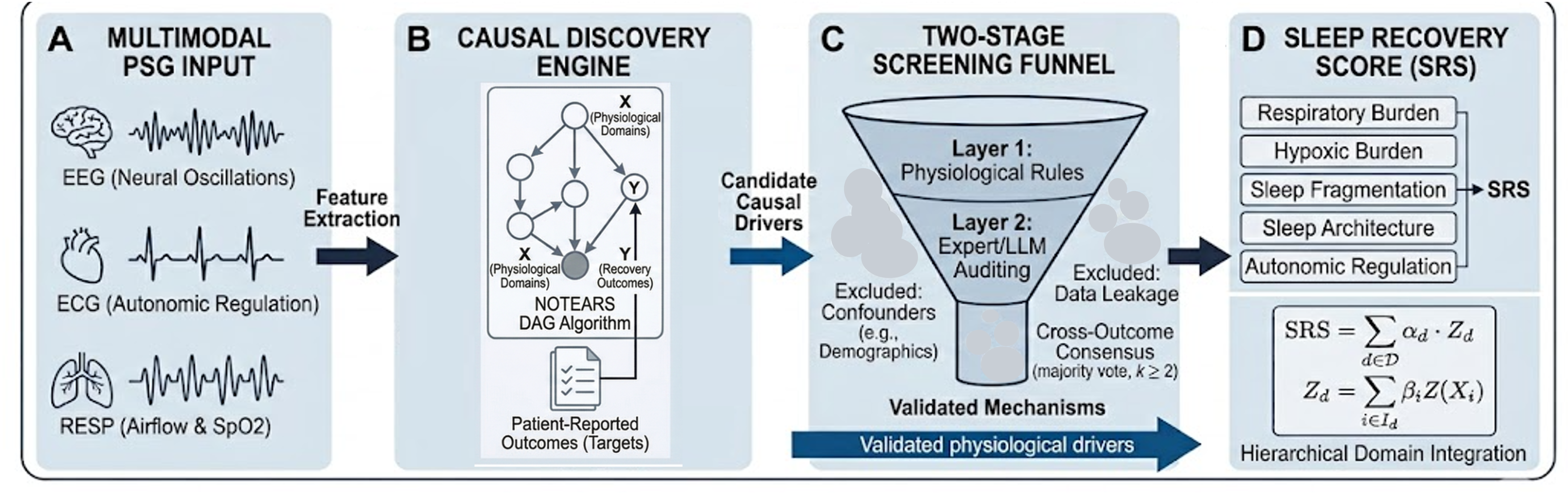}
  \caption{
Overview of the proposed sleep recovery framework.
(A) Multimodal polysomnography (PSG) signals, including
EEG, ECG, and respiratory/oximetry channels, are
converted into physiological features.
(B) A NOTEARS-based directed acyclic graph (DAG) model
identifies candidate physiological drivers of
recovery-related outcomes.
(C) A two-stage screening funnel applies physiology-based
constraints and constrained LLM-assisted auditing to remove
confounders and construct-overlapping variables.
(D) Screened mechanisms are aggregated into five
physiological domains---respiratory burden, hypoxic burden,
sleep fragmentation, sleep architecture, and autonomic
regulation---to derive the hierarchical Sleep Recovery
Score (SRS).
}
  \label{fig:srs_framework}
\end{figure*}
\section{Introduction}

Connected health systems increasingly incorporate multimodal
physiological sensing---including wearable electrocardiography
(ECG), pulse oximetry, and consumer sleep devices---yet
translating continuous biosignals into clinically interpretable
recovery metrics remains an open challenge. A central difficulty
is that distributed sensing platforms often capture only partial
views of sleep physiology, making domain-structured and
interpretable recovery modeling especially important for
translation beyond the sleep laboratory. Polysomnography
(PSG), the clinical gold standard for sleep assessment, captures
rich multimodal recordings spanning electroencephalography (EEG),
ECG, respiratory flow, and oxygen saturation~\cite{Berry2012AASM}.
Despite this richness, clinical practice often reduces these
signals to coarse summary indices such as the Apnea--Hypopnea
Index (AHI) or sleep efficiency. While clinically useful, these
measures primarily capture isolated components of sleep
disruption and may fail to reflect the integrated physiological
burden underlying functional recovery~\cite{Azarbarzin2019HypoxicBurdenCVD,
Peker2025HypoxicBurdenVsAHI}.

Patient-reported outcomes (PROs), including daytime sleepiness
and fatigue, capture the lived consequences of sleep dysfunction
but do not directly reveal the physiological systems driving
impaired recovery. Prior studies have reported only
weak-to-moderate alignment between respiratory event metrics and
subjective symptom burden~\cite{Johns1991ESS,Guo2020WeightedESS},
highlighting a persistent disconnect between objective sleep
physiology and perceived recovery. At the same time, recent
machine learning approaches for PSG analysis have improved
predictive performance, but often operate as black-box systems
optimized for association rather than mechanistic
insight~\cite{Perslev2021USleep,Thapa2026SleepFM}. In
sleep medicine and connected health, however, interpretability,
physiological plausibility, and robustness are essential for
clinical trust and responsible deployment.

To address this gap, we propose an interpretable,
causal-discovery--guided framework for deriving a hierarchical
Sleep Recovery Score (SRS) from multimodal PSG. Rather than
relying on heuristic aggregation or single-index severity
measures, we treat recovery-related PROs as structured targets
and use directed acyclic graph (DAG) modeling to identify
candidate physiological drivers of recovery. These mechanisms are
then refined through a two-stage screening process that combines
physiology-based constraints with constrained LLM-assisted
auditing, and are aggregated into five physiological domains:
respiratory burden, hypoxic burden, sleep fragmentation, sleep
architecture, and autonomic regulation. Although derived from
clinical PSG, these domains align naturally with sensing streams
increasingly available in connected health technologies,
including wearable ECG, oximetry, and sleep-stage estimation
devices. We evaluate the proposed framework in two independent
population cohorts, MESA and MrOS. Across cohorts, the resulting
SRS shows stronger alignment with perceived recovery than AHI,
with improvements of up to 3.4$\times$ for some outcomes. By
explicitly linking multimodal sleep physiology to patient-centered
recovery through a transparent and domain-structured framework,
this work provides a practical foundation for recovery modeling
across both clinical sleep studies and emerging smart and
connected health settings.

\noindent\textbf{Contributions.}
\begin{enumerate}
    \item \textbf{Interpretable multidomain recovery modeling.}
    We introduce a hierarchical Sleep Recovery Score that moves
    beyond single respiratory indices by linking multimodal PSG
    to patient-reported recovery outcomes.

    \item \textbf{Causal-discovery--guided mechanism identification.}
    We apply DAG-based causal discovery to identify candidate
    physiological drivers of recovery, refined through
    physiology-based constraints and constrained LLM-assisted
    auditing.

    \item \textbf{Cross-cohort evidence supporting connected-health translation.}
    Across MESA and MrOS, we observe consistent convergence on
    five physiological domains and show stronger alignment with
    perceived recovery than AHI, supporting interpretable
    recovery modeling that can inform future distributed and
    wearable sensing systems.
\end{enumerate}

\section{Related Work}

Clinical sleep assessment remains dominated by summary indices
such as the Apnea--Hypopnea Index (AHI), despite growing
evidence that single respiratory metrics do not fully capture
downstream physiological burden or functional
recovery~\cite{Berry2012AASM,Azarbarzin2019HypoxicBurdenCVD,
Peker2025HypoxicBurdenVsAHI}. Patient-reported outcomes such as
daytime sleepiness provide an important complementary view of
sleep dysfunction, yet prior work has reported only
weak-to-moderate alignment between subjective symptom burden and
PSG-derived respiratory indices~\cite{Johns1991ESS,
Guo2020WeightedESS}. These findings motivate multidomain
recovery models that move beyond event-count severity alone.

Recent machine learning methods have improved sleep staging and
PSG-based prediction from multimodal physiological signals,
including deep learning and foundation-model
approaches~\cite{Perslev2021USleep,Thapa2026SleepFM}. However,
these systems typically optimize predictive performance rather
than mechanistic interpretability. Causal discovery methods such
as NOTEARS provide a complementary approach for identifying
structured dependencies in observational
data~\cite{Zheng2018NOTEARS}, but in clinical settings, learned
graphs still require mechanism-aware screening to avoid
confounders, proxy effects, and physiologically implausible
relationships. Related graph-based work on invasive neural
recordings has shown how interpretable models can expose dynamic,
behavior-conditioned structure in physiological time series.
Asadi developed graph-based representations of deep-brain dynamics
during motor behavior, and the BACE framework extended this idea to
phase-specific directed connectivity estimation from multiregion
local field potentials~\cite{Asadi2025DeepBrain,Asadi2025BACE}.
Although these studies address neural rather than sleep physiology,
they motivate transparent, context-sensitive graph representations.
In contrast to prior PSG-based predictive
models, our goal is not only outcome association but extraction
of recurrent, physiologically interpretable recovery domains that
can support downstream connected-health translation. Our work
builds on these directions by combining multimodal PSG,
causal-discovery--guided mechanism identification, and
structured candidate screening to derive an interpretable
recovery score for connected health systems.
\section{Methodology}
\label{sec:methodology}

We derive the Sleep Recovery Score by linking PSG features to recovery-related
patient-reported outcomes through a structured
causal-discovery pipeline. As shown in
Figure~\ref{fig:srs_framework}, the framework consists of five
steps: (1) outcome selection, (2) physiological feature
construction, (3) directed acyclic graph (DAG) estimation,
(4) two-stage candidate screening, and (5) hierarchical score
synthesis. We apply this framework to two population cohorts,
MESA ($n=1{,}540$) and MrOS ($n=825$), using recovery outcomes
that include daytime sleepiness, fatigue, perceived sleep
quality, and perceived sleep efficiency.

\subsection{Causal Graph Estimation}

Let $X \in \mathbb{R}^{n \times p}$ denote the matrix of $p$
candidate physiological features measured across $n$
individuals, and let $Y_k$ denote a recovery-related PRO.
Features span five domains of sleep physiology: respiratory
burden, hypoxic burden, sleep fragmentation, sleep
architecture, and autonomic regulation. Representative
variables include AHI and hypopnea index for respiratory
burden, SpO$_2$ and oxygen desaturation index (ODI) for
hypoxic burden, wake after sleep onset (WASO) and arousal
index for sleep fragmentation, N3\%, REM latency, and spectral
power for sleep architecture, and SDNN and RMSSD for
autonomic regulation.

For each outcome $Y_k$, we form an analysis table
$T_k = [X, C, Y_k]$, where $C$ contains structural covariates
such as age, sex, race, and education. Continuous variables are
median-imputed and standardized, while categorical covariates
are one-hot encoded. NOTEARS was run with sparsity parameter
$\lambda_1 = 0.02$ and edge-weight threshold $\tau = 0.01$,
retaining the top-$k = 20$ candidate drivers per outcome
(Fisher-Z independence test, $\alpha = 0.05$). Bootstrap
stability selection used 500 resamples; only edges with
selection frequency $\geq 0.6$ were carried forward to the
screening stage.

We estimate a sparse DAG over the variables in $T_k$ using the
linear NOTEARS formulation~\cite{Zheng2018NOTEARS}:
\begin{equation}
\begin{aligned}
\min_{W} \quad
& \frac{1}{2n}\|X - XW\|_F^2 + \lambda_1 \|W\|_1 \\
\text{subject to} \quad
& \mathrm{tr}\!\left(e^{W \circ W}\right) - d = 0
\end{aligned}
\label{eq:notears}
\end{equation}
where $W \in \mathbb{R}^{d \times d}$ is the weighted adjacency
matrix, $\lambda_1$ controls sparsity, and $d$ is the number of
modeled variables. We adopt the linear formulation to
prioritize interpretability and stability in moderate-sample
biomedical settings, yielding an explicit sparse dependency
structure that can be inspected and audited against physiology
before score construction.

Features with directed edges into $Y_k$ are treated as
candidate drivers of recovery. To improve robustness, we apply
bootstrap stability selection and retain only edges whose
selection frequency exceeds a threshold $\tau$, yielding a
sparse and reliability-filtered candidate set.

\textit{Interpretive note on causal language.}
Although we adopt causal-discovery terminology throughout, MESA
and MrOS are observational cross-sectional cohorts. Learned DAG
edges therefore reflect conditional statistical dependencies
rather than established causal relationships. The two-stage
screening pipeline imposes domain-motivated plausibility
constraints that go beyond statistical structure, but cannot
substitute for experimental validation. We use the term
\textit{candidate driver} deliberately to signal a mechanism
hypothesis rather than a causal claim, consistent with the
exploratory role of causal discovery in biomedical observational
settings~\cite{Zheng2018NOTEARS}.

\subsection{Two-Stage Candidate Screening}

Because causal discovery may recover statistically valid but
physiologically implausible or clinically unhelpful
relationships, we refine candidate drivers through a two-stage
screening process.

\textbf{Stage 1: Physiology-based screening.}
Candidate edges are filtered using established sleep
mechanisms. Representative pathways include
OSA $\rightarrow$ hypoxia $\rightarrow$ arousals
$\rightarrow$ sleepiness, reduced N3 activity
$\rightarrow$ impaired restoration, and elevated alpha/beta
activity $\rightarrow$ cortical hyperarousal. Edges
inconsistent with known physiology are removed.

\textbf{Stage 2: Constrained LLM-assisted auditing.}
Remaining candidates are submitted to a structured three-way
classification protocol implemented in
\texttt{validate\_with\_llm.py} (see Code Availability)
using a constrained LLM-assisted audit protocol. Each candidate feature is classified as:
(i)~\textit{plausible mechanistic driver},
(ii)~\textit{structural confounder} (e.g., demographic proxy
such as race, sex, or age), or
(iii)~\textit{construct-overlapping leakage variable}
(e.g., a subjective measure correlated by construction with
the PRO target). Classification is governed by predefined
physiological and methodological criteria supplied as fixed
system-level instructions, ensuring consistent and reproducible
auditing across all outcomes. Only features classified as
category~(i) by both layers advance to consensus aggregation.
Full prompt templates and classification criteria are included
in the repository.

Across multiple recovery outcomes $\{Y_k\}_{k=1}^{K}$, the final
retained mechanism set $\mathcal{M}$ is obtained by
cross-outcome consensus aggregation, retaining features that
recur in at least two outcomes ($k \geq 2$). This majority-vote
step reduces outcome-specific noise and improves robustness.

\subsection{Hierarchical SRS Construction}
\label{subsec:srs}

Retained features are grouped into five physiological domains:
respiratory burden, hypoxic burden, sleep fragmentation, sleep
architecture, and autonomic regulation. For each domain $d$,
let $\mathcal{I}_d$ denote the subset of retained features
assigned to that domain. We compute the domain score as
\begin{equation}
Z_d = \sum_{j \in \mathcal{I}_d} \beta_j\, Z(X_j),
\label{eq:domain}
\end{equation}
where $\beta_j$ is proportional to NOTEARS edge magnitude and
cross-outcome stability frequency, and is normalized such that
$\sum_{j \in \mathcal{I}_d} |\beta_j| = 1$. Feature signs are
oriented so that higher values of the final score correspond to
better physiological recovery.

The overall Sleep Recovery Score is then defined by
hierarchical aggregation across domains:
\begin{equation}
\mathrm{SRS} = \sum_{d} \alpha_d\, Z_d,
\label{eq:srs}
\end{equation}
where $\alpha_d$ denotes the domain-level weight. Domain
weights are set proportional to the mean cross-outcome
stability of retained features within each domain and are
normalized so that $\sum_d |\alpha_d| = 1$.

This domain-structured formulation also supports partial score
estimation when only subsets of sensing channels are available,
which is important for translation to distributed and wearable
monitoring settings.

Finally, the score is standardized within cohort:
\begin{equation}
\mathrm{SRS}_{\mathrm{final}}^{(c)}
=
\frac{\mathrm{SRS}^{(c)} - \mu_c}{\sigma_c},
\label{eq:srs_std}
\end{equation}
where $\mu_c$ and $\sigma_c$ denote the cohort-specific mean
and standard deviation. Higher values of
$\mathrm{SRS}_{\mathrm{final}}$ therefore indicate greater
multidomain physiological recovery.
\begin{table*}[t]
\centering
\renewcommand{\arraystretch}{1.15}
\caption{Physiological domains underlying the Sleep Recovery Score (SRS). Representative drivers retained after causal discovery, two-stage candidate screening, and cross-outcome consensus in MESA and MrOS.}
\label{tab:mechanistic_convergence}

\begin{tabular}{p{3.7cm} p{8.8cm}}
\toprule
\textbf{Physiological Domain} & \textbf{Representative Drivers} \\ 
\midrule

Respiratory Burden 
& Apnea--Hypopnea Index (AHI), hypopnea index, total respiratory events \\

Hypoxic Burden 
& Oxygen desaturation index (ODI3/ODI4), T90, minimum and mean nocturnal SpO$_2$ \\

Sleep Fragmentation 
& Wake after sleep onset (WASO), arousal index, stage transitions \\

Sleep Architecture / EEG 
& Reduced N3, altered REM proportion and latency, elevated alpha/beta spectral power \\

Autonomic Regulation 
& Heart rate variability (SDNN, RMSSD), elevated nocturnal heart rate \\

\bottomrule
\end{tabular}

\vspace{4pt}
\footnotesize
\textit{Domains were retained if they appeared in at least two recovery-related outcomes ($k \geq 2$) and were retained after physiology-based screening and constrained LLM-assisted auditing.}
\end{table*}
\begin{figure}[t]
\centering
\includegraphics[width=\linewidth]{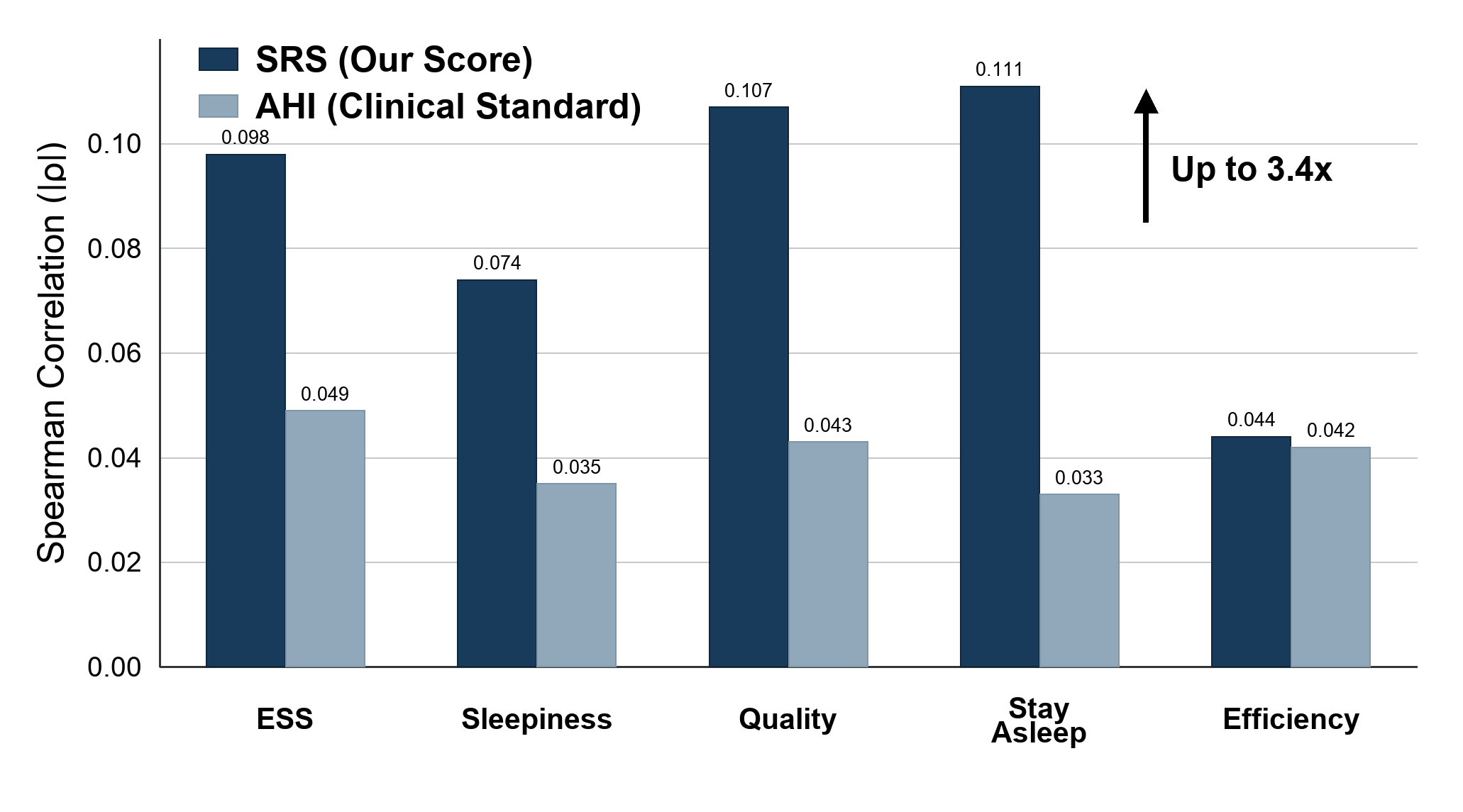}
\caption{SRS aligns more strongly with perceived recovery than AHI. Spearman correlations ($\rho$) between the proposed Sleep
Recovery Score (SRS) and the Apnea--Hypopnea Index (AHI) across
recovery-related outcomes in MESA and MrOS. In both cohorts,
SRS shows consistently stronger associations with perceived
recovery, including up to a 3.4$\times$ improvement for trouble
staying asleep in MrOS.}
\label{fig:perf}
\end{figure}

\section{Results}
\label{sec:results}

\subsection{Feature Distillation and Mechanism Screening}

We performed structured feature distillation to reduce the full
physiological feature space to the final retained mechanisms
used for SRS construction.

In MESA, 53 physiological features were initially evaluated.
Linear NOTEARS identified a median of 44 candidate drivers
across recovery outcomes. Physiology-based screening reduced
this set to 34 biologically plausible mechanisms. After
constrained LLM-assisted auditing and removal of structural
covariates and construct-overlapping variables, 32 consensus
mechanisms were retained for hierarchical score construction.

In MrOS, 174 physiological features were initially considered.
DAG estimation identified a median of 58 candidate drivers per
outcome. Physiology-based screening reduced this set to 30
plausible mechanisms, and cross-outcome consensus aggregation
yielded 28 final retained mechanisms.

Structural covariates such as age, sex, and race were
occasionally identified as statistical parents of recovery
outcomes, but were excluded during screening to prevent proxy
effects and preserve mechanistic interpretability. In MrOS,
sleep latency and wake after sleep onset (WASO) were identified
as construct-overlapping leakage variables for subjective sleep
quality and were removed from final aggregation. These results
show that SRS is not a heuristic combination of PSG variables,
but the outcome of causal discovery, physiology-based
screening, and bias-aware candidate filtering.

\subsection{Cross-Cohort Mechanistic Convergence}

Across both cohorts and heterogeneous recovery instruments, five physiological domains consistently emerged as
recurrent physiological correlates of impaired recovery:
(i) respiratory burden,
(ii) hypoxic burden,
(iii) sleep fragmentation,
(iv) sleep architecture and EEG dynamics, and
(v) autonomic regulation.

Table~\ref{tab:mechanistic_convergence} summarizes
representative retained drivers within each domain. These
domains capture complementary aspects of impaired recovery,
spanning disordered breathing, oxygenation, fragmentation,
neurophysiological architecture, and autonomic regulation.

The recurrence of these five domains across MESA and MrOS
supports the biological robustness of the framework and reduces
the likelihood that the resulting score is driven by
cohort-specific statistical artifacts.
\begin{table*}[t]
\centering
\renewcommand{\arraystretch}{1.2}
\caption{Spearman $\rho$ with 95\% CIs (Fisher $z$-transformation) for SRS and AHI across recovery outcomes (MESA: $n=1{,}540$; MrOS: $n=825$). $^{***}p{<}0.001$, $^{**}p{<}0.01$, $^{\mathrm{ns}}p{>}0.05$.}
\label{tab:spearman}
\begin{tabular}{llp{4.2cm}p{4.2cm}}
\toprule
\textbf{Outcome} & \textbf{Cohort} & \textbf{SRS $\rho$ [95\% CI]} & \textbf{AHI $\rho$ [95\% CI]} \\
\midrule
ESS         & MESA & $0.098\ [0.048,\ 0.147]^{***}$         & $0.049\ [-0.001,\ 0.099]^{\mathrm{ns}}$ \\
Sleepiness  & MESA & $0.074\ [0.024,\ 0.123]^{**}$          & $0.035\ [-0.015,\ 0.085]^{\mathrm{ns}}$ \\
Quality     & MrOS & $0.107\ [0.039,\ 0.174]^{**}$          & $0.043\ [-0.025,\ 0.111]^{\mathrm{ns}}$ \\
Stay Asleep & MrOS & $0.111\ [0.043,\ 0.178]^{**}$          & $0.033\ [-0.035,\ 0.101]^{\mathrm{ns}}$ \\
Efficiency  & MrOS & $0.044\ [-0.024,\ 0.112]^{\mathrm{ns}}$ & $0.042\ [-0.026,\ 0.110]^{\mathrm{ns}}$ \\
\bottomrule
\end{tabular}
\end{table*}

\subsection{Association with Recovery Outcomes}

Table~\ref{tab:spearman} reports Spearman $\rho$ with 95\%
confidence intervals and $p$-values for all outcomes
(MESA: $n=1{,}540$; MrOS: $n=825$; CIs via Fisher
$z$-transformation). SRS was statistically significant for
four of five outcomes, while AHI reached significance for
none. In MESA, SRS showed significant associations with
ESS ($\rho = 0.098$, 95\%~CI $[0.048, 0.147]$, $p < 0.001$)
and daytime sleepiness ($\rho = 0.074$, $[0.024, 0.123]$,
$p = 0.004$), whereas AHI was non-significant for both
(ESS: $p = 0.055$; sleepiness: $p = 0.170$). In MrOS, SRS
showed significant associations with perceived sleep quality
($\rho = 0.107$, $[0.039, 0.174]$, $p = 0.002$) and trouble
staying asleep ($\rho = 0.111$, $[0.043, 0.178]$,
$p = 0.001$), while AHI was non-significant for both
($p = 0.217$ and $p = 0.344$, respectively). For perceived
sleep efficiency, neither SRS nor AHI reached significance
($p = 0.207$ and $p = 0.228$), consistent with the
well-documented difficulty of predicting subjective efficiency
from objective physiology alone.

These findings suggest that restorative sleep is better
characterized as a multidomain physiological construct than
as a single respiratory index alone. Rather than replacing
disease-specific metrics such as AHI, SRS complements them
by capturing the broader physiological substrate of recovery.

\section{Discussion}
\label{sec:discussion}

This work reframes sleep recovery modeling from single-index
severity estimation toward structured, multidomain
physiological integration. We introduced a
causal-discovery--guided framework for constructing an
interpretable and hierarchical Sleep Recovery Score (SRS) from
polysomnography (PSG) and patient-reported outcomes. Across two
independent cohorts (MESA and MrOS), the framework converged on
a stable set of physiologically coherent domains underlying
impaired recovery.

\subsection{Mechanistic Convergence Across Cohorts}
Across both cohorts, the framework converged on five
physiological domains consistently associated with impaired recovery:
respiratory burden, hypoxic burden, sleep fragmentation, sleep
architecture and EEG dynamics, and autonomic regulation
(Table~\ref{tab:mechanistic_convergence}).

The recurrence of these domains across two independent cohorts
suggests that impaired recovery is associated with a stable multidomain
physiological structure rather than dataset-specific variation.
Instead of collapsing heterogeneous mechanisms into a single
severity metric, the hierarchical SRS preserves domain-level
structure and enables transparent decomposition of respiratory,
hypoxic, neurophysiological, fragmentation, and autonomic
contributions to impaired recovery.

\subsection{Interpretability and Structural Validity}

Unlike purely predictive models, the proposed framework
explicitly distinguishes mechanistic physiological drivers from
structural covariates and construct-overlapping variables.
Demographic features such as age, sex, and race were
occasionally identified as statistical parents during graph
estimation, but were excluded from SRS construction to prevent
proxy effects. Similarly, variables overlapping with subjective
constructs were removed during candidate screening.

This structured filtering helps ensure that SRS reflects
physiological mechanisms rather than demographic or measurement
artifacts, strengthening interpretability and structural
validity.

\subsection{Performance Relative to Single-Index Metrics}

SRS consistently demonstrated stronger and statistically more
reliable alignment with patient-reported outcomes than AHI
across both cohorts (Table~\ref{tab:spearman}). Notably, AHI
failed to reach significance for all five outcomes, whereas
SRS was significant for four of five---suggesting that
multidomain integration provides not merely stronger but
qualitatively more reliable associations with perceived
recovery. For perceived sleep efficiency, the non-significance
of both metrics reflects the complex, multifactorial nature of
subjective efficiency as a construct, representing an honest
boundary of the framework's current performance.

These findings support the view that restorative sleep reflects
distributed multidomain physiological burden rather than a
single dominant respiratory index. Rather than replacing
disease-specific measures such as AHI, SRS complements them by
capturing the broader physiological substrate of recovery.

\subsection{Clinical and Connected Health Implications}

The hierarchical structure of SRS provides more than a
composite severity score; it offers a mechanism-aware framework
for targeted interpretation and a natural pathway toward
translation in connected health systems. Because SRS is
constructed from distinct physiological domains, domain-level
contributions ($Z_d$) can be inspected to identify the dominant
driver of impaired recovery in an individual patient.

For example, recovery impairment driven primarily by hypoxic
burden may motivate airway-focused interventions, whereas
impairment dominated by autonomic dysregulation may motivate
cardiovascular or behavioral strategies. Traditional AHI-based
assessment cannot distinguish between these mechanistic
profiles.

Beyond laboratory polysomnography, the domain-structured
formulation enables modular scalability. Wearable ECG,
oximetry, and sleep staging devices capture subsets of these
domains, allowing partial but interpretable recovery estimation
in distributed sensing environments. This makes SRS a promising
foundation for smart and connected health systems.

\subsection{Limitations}

Several limitations warrant consideration. First, the linear
NOTEARS formulation may not capture nonlinear physiological
interactions. Second, causal structure learning relies on
assumptions of causal sufficiency and may omit latent
variables. Third, correlations with subjective outcomes were
modest, reflecting the multifactorial nature of perceived
recovery. Fourth, the second-stage auditing process relied on a
constrained LLM-based procedure rather than domain-expert
review, which may affect the reliability of candidate
classification. Finally, the analysis was cross-sectional and
does not establish temporal causality. Future work should
explore nonlinear causal discovery, expert-reviewed or
prospectively evaluated mechanism screening, longitudinal
validation, and intervention-sensitive recovery modeling.

\section{Conclusion}
\label{sec:conclusion}

We introduced a causal-discovery--guided framework for
constructing an interpretable, hierarchical Sleep Recovery Score
from polysomnography and patient-reported outcomes. Across two
independent cohorts, the framework converged on five
physiologically coherent domains and showed stronger alignment
with perceived recovery than AHI. These findings support a
multidomain view of sleep recovery and suggest that structured,
mechanism-aware modeling can provide a practical foundation for
interpretable recovery assessment in smart and connected health
systems.

\bibliographystyle{IEEEtran}
\bibliography{references}

\end{document}